\documentclass[11pt]{article}

\usepackage[preprint]{acl}
\usepackage{times}
\usepackage{latexsym}

\usepackage[T1]{fontenc}

\usepackage[utf8]{inputenc}

\usepackage{microtype}

\usepackage{inconsolata}
\usepackage{graphicx}
\usepackage{amsmath}
\usepackage{amssymb}
\usepackage{xcolor}
\usepackage{hyperref}
\title{HRGraph: Leveraging LLMs for HR Data Knowledge Graphs with Information Propagation-based Job Recommendation}

\author{
 \textbf{Azmine Toushik Wasi}
\\
Shahjalal University of Science and Technology, Bangladesh
\\
\texttt{azmine32@student.sust.edu}
}

\begin{document}
\maketitle
\begin{abstract}
Knowledge Graphs (KGs) serving as semantic networks, prove highly effective in managing complex interconnected data in different domains, by offering a unified, contextualized, and structured representation with flexibility that allows for easy adaptation to evolving knowledge. Processing complex Human Resources (HR) data, KGs can help in different HR functions like recruitment, job matching, identifying learning gaps, and enhancing employee retention. Despite their potential, limited efforts have been made to implement practical HR knowledge graphs. This study addresses this gap by presenting a framework for effectively developing HR knowledge graphs from documents using Large Language Models. The resulting KG can be used for a variety of downstream tasks, including job matching, identifying employee skill gaps, and many more. In this work, we showcase instances where HR KGs prove instrumental in precise job matching, yielding advantages for both employers and employees. Empirical evidence from experiments with information propagation in KGs and Graph Neural Nets, along with case studies underscores the effectiveness of KGs in tasks such as job and employee recommendations and job area classification. Code and data are available at : \href{https://github.com/azminewasi/HRGraph}{https://github.com/azminewasi/HRGraph}
\end{abstract}



\section{Introduction}
Knowledge Graph (KG) is a semantic network that stores real-world entities and their relationships. It uses nodes representing objects, places, or persons, connected by edges defining relationships. It can integrate diverse data, contextualize information through linking and semantic metadata, and remain flexible, accommodating dynamic knowledge changes seamlessly \cite{Hogan_2021, wasi-etal-2024-banglaautokg-automatic,yang2024common,khorashadizadeh2023exploring}.



\begin{figure}[t] 
\centering {
\includegraphics[scale=.52]{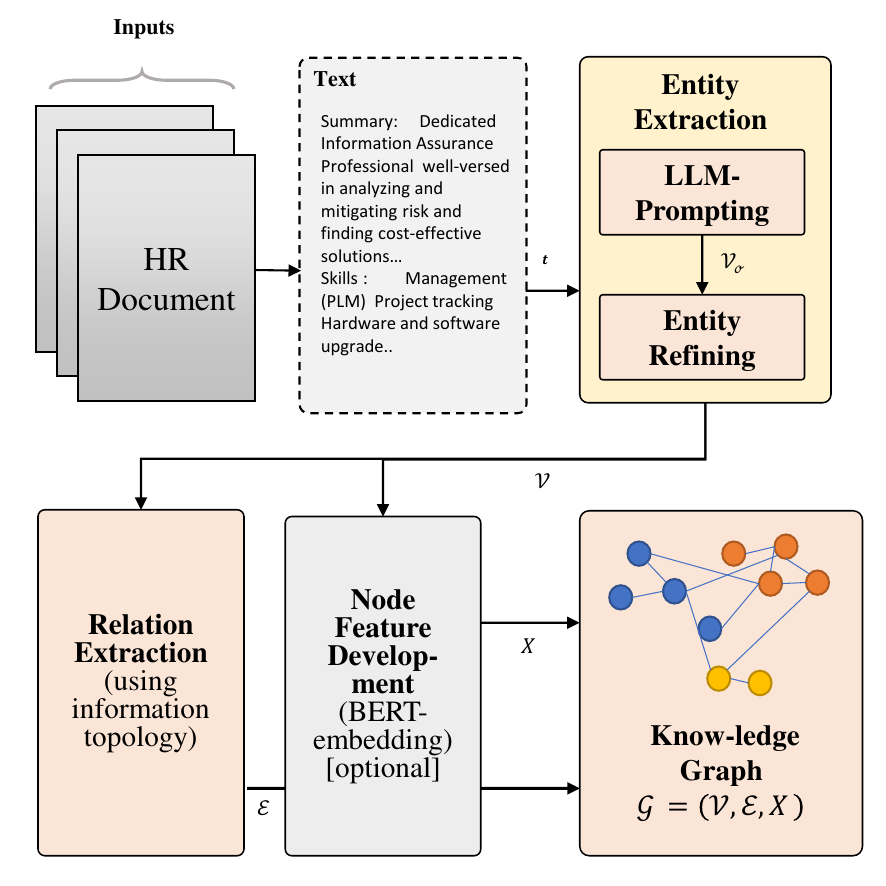}
\vspace{-6mm}
\caption{The overall framework of our \textbf{HRGraph}. It involves passing text data extracted from HR documents through a LLM to obtain entities and entity types, which are used to build a base knowledge graph with optional node features as BERT embeddings.}
}\label{fig:mainHRKG}
\vspace{-6mm}
\end{figure}

Knowledge Graphs can be highly effective for managing HR data, integrating diverse sources into a unified, structured representation \cite{9515238efwfg, wasi-etal-2024-banglaautokg-automatic}. This is crucial for applications like recruitment and career path planning. By linking data with semantic metadata, KGs prevent misinterpretation, particularly in employee skill mapping and development. Their flexibility allows easy adaptation to new data and requirements across various HR functions. KGs enhance recruitment precision, skill and career mapping, optimize recruitment processes, identify learning gaps, improve retention strategies, and facilitate organizational knowledge sharing. For employees, KGs offer better job searches and recommendations, providing strong support from their perspective \cite{10187afeaf510,Bao2dsggdrsg021}.


\begin{figure*}[t] 
\centering {\includegraphics[scale=0.25]{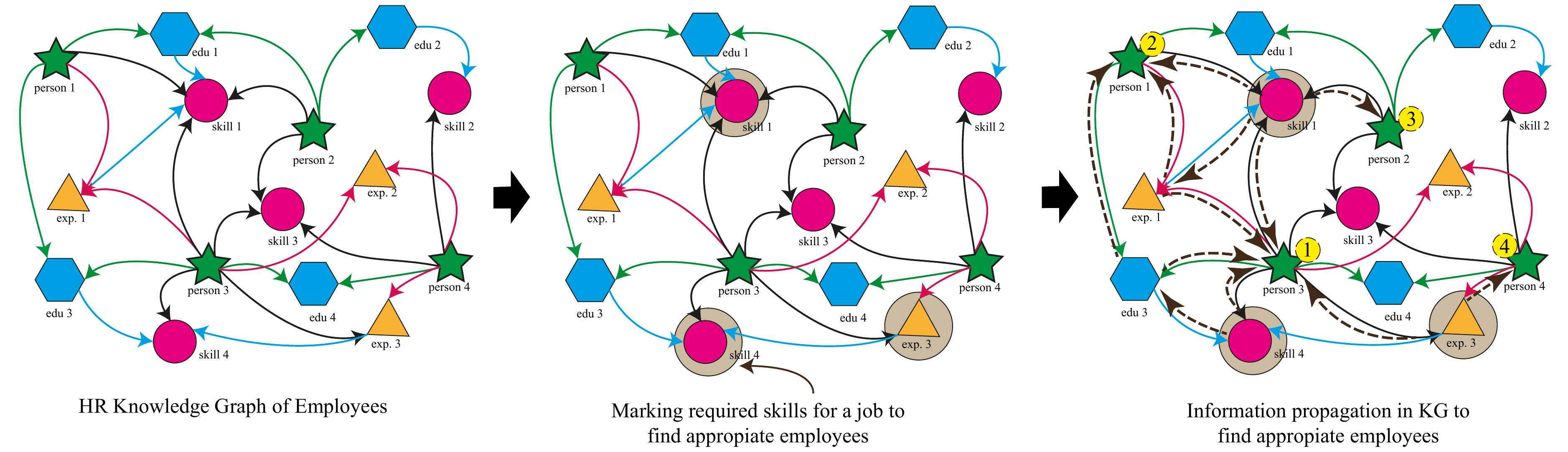}}
\vspace{-2mm}
\caption{An overview of Job Matching and Recommendation framework using HR KGs of job applicants or employees. Using the intuition that company JDs and employee CVs or profiles should share matching entities like skills, experience, and education, we can match, find, and recommend employees from a knowledge graph. On the other hand, using a KG of many different job descriptions, we can use job-seeker skills, education, and other factors to find an appropriate employment to recommend.}\label{fig:what-m}
\vspace{-6mm}
\end{figure*}

In this study, we introduce a framework named \textbf{HRGraph}, aimed at constructing HR knowledge graphs from various HR documents, such as Job Descriptions (JDs) and Curriculum Vitae (CVs). We illustrate the framework's utility through two practical examples as downstream tasks: employee and job recommendation using the HR knowledge graph. Our approach employs Large Language Models (LLMs) to identify and extract diverse entities, then extracts node features using pre-trained BERT, forming the knowledge graph with some post-processing. The resulting knowledge graph can be utilized for different downstream tasks. 
In this work, we use it for effective job matching and classification, catering to both employer and employee needs. The underlying idea is that company JDs and employee CVs or profiles should share matching entities like skills, experience, and education (an illustration of the proposal is presented in Figure \ref{fig:what-m}), facilitating a comprehensive and accurate job match in both job-seeking and employee-search scenarios.

\section{Related Works}
\subsection{Applications of Knowledge Graphs}
KGs showcase versatility, excelling in applications such as semantic search, question answering, and recommendation systems \citep{wang2023knowledge, gao2020deep, wasi-etal-2024-banglaautokg-automatic}. Their structured representation enhances search engine results and tailors suggestions. KGs, pivotal in Natural Language Processing (NLP), elevate information extraction and contribute to superior machine learning predictions. From enterprise knowledge management to biomedical research \citep{10026520}, KGs exhibit adaptability. Their integration of diverse data, contextualization, and inherent flexibility underpin effectiveness in managing and extracting insights across varied domains, including Medical AI \citep{10026520}, Wireless Communication Networks \citep{he2022representation}, Search Engines \citep{heist2020knowledge}, and Big Data Decision Analysis \citep{2020KGbook}. KGs emerge as indispensable tools, navigating dynamic information landscapes seamlessly \citep{Hogan_2021}. Our work is inspired by these different uses of knowledge graphs.

\subsection{HR Data and Knowledge Graphs}
Though human resource knowledge graphs have good potential, limited efforts have been made in this domain. \citet{9515238efwfg} adopted a top-down approach to create the ontology model of a human resource knowledge graph. The paper describes the significance of initially establishing ontology and defines entities and relationships for HR KGs. {\citet{khhbfrahbir} presented a hypothesis to build a job description KG using NLP-based semantic entity extraction, but no detailed methodologies or experiments were presented. \citet{10065776ewe} presented a job recommendation algorithm based on KGs, using word similarity to find recommendations. \citet{965875asegasg7} uses a NER-based approach to build knowledge graphs to aid in job recommendation.} However, no practical efforts have been made to implement a tangible HR knowledge graph in real-world scenarios based on LLMs and utilize one graph for multiple downstream tasks.

\section{Methodology}
The recent developments in LLMs, Knowledge Graph-based systems, and GNNs served as inspiration for the proposed methodology. Inspired by \citet{wasi-etal-2024-banglaautokg-automatic}, HR knowledge graph uses Large Language Models (LLMs) for entity extraction and pre-trained NLP moels for node features enables a sophisticated representation of HR data by leveraging advanced language understanding capabilities. 
Existing literature presents alternative approaches, such as  word similarity-based job recommendation \citep{10065776ewe} and NERs to build KGs \citep{965875asegasg7}. Our proposed method stands out by leveraging LLMs for entity extraction and using BERT for features (length 256), offering a flexible and comprehensive approach that addresses practical challenges and enables multiple downstream tasks.

\textbf{Entity Extraction and Refining.}
We begin by processing a job description (JD) or curriculum vitae (CV) as HR document text $t$, using a Large Language Model Gemini \cite{geminiteam2023gemini}. Prompts are available in Section \ref{sec:prompts}. This step results in the extraction of various entities ($\mathbb{V}^o$) from the text, capturing both entities and relationships. Subsequently, we perform post-processing on the entity set ($\mathbb{V}^o$) to filter out potential noises (such as KG nodes having more than 3 words or having no named entity or verbs), resulting in a refined set of entities $\mathbb{V}$.

\textbf{Relation Extraction.}
To establish the initial connections between these entities, we leverage information topology and types, creating the initial connections set $\mathcal{E}$. 

\textbf{Node Feature Extraction.}
Employing a pre-trained BERT model, we generate feature vectors for each entity, constructing an initial feature matrix $X$. Thus, $\mathcal{V}$ represents the ensemble of nodes (entities) $\{v_1, v_2, v_3, \cdots v_N\}$, and $\mathcal{E}$ encompasses the collection of edges (relationships) $\{e_1, e_2, e_3, \cdots e_M\}$, where $N$ and $M$ signify the number of nodes and edges, respectively.

\textbf{Knowledge Graph Construction.}
Combining $\mathcal{V}$, $\mathcal{E}$, and $X$ forms our Knowledge Graph (KG), denoted as $\mathcal{G}^o = (\mathcal{V}, \mathcal{E}, X)$. The corresponding adjacency matrix, $\mathcal{A}$, has an element $\mathcal{A}_{ij}=1$ if an edge connects $v_i$ and $v_j$.

Each node $v \in \mathcal{V}$ and each edge $e \in \mathcal{E}$ have associated mapping functions, denoted as $\phi(v): \mathcal{V} \rightarrow \mathcal{A}$ and $\varphi(e): \mathcal{E} \rightarrow \mathcal{R}$. Here, $\mathcal{R}$ represents the edge type set, and $\mathcal{A}$ is the node type set, where $|\mathcal{A}|+|\mathcal{R}|>2$. If we choose to use Knowledge Graph Embedding (KGE) \cite{cao2023knowledge}, feature vector $X$ can be excluded, and node embeddings can be obtained using different KGE models containing topological and structural knowledge.

\begin{figure}[t] 
\centering {\includegraphics[scale=0.22]{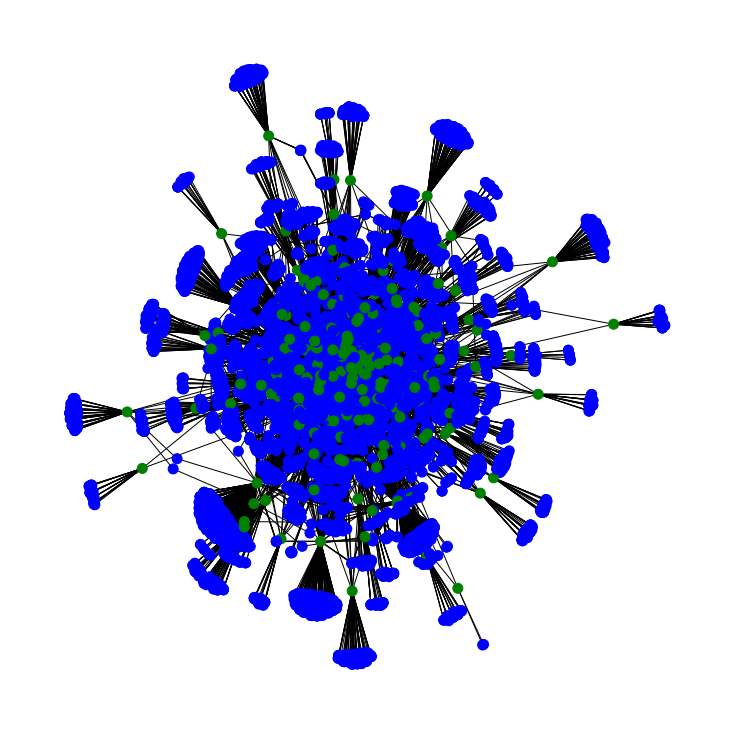}}
\vspace{-4mm}
\caption{Our CV Knowledge Graph. Green dots are persons (CV), blue dots are skills, education and other entities}\label{fig:KG_Examples/CV_mega}
\vspace{-4mm}
\end{figure}

\section{Experiments with Knowledge Graphs}
We collect 200 CVs and 200 job descriptions from online job portals, ensuring the CVs had minimal personal information and underwent manual review for privacy protection. Company details, openly available in job descriptions, were retained. The data was then manually labeled with the help of job portal filters for subsequent experiments. Any type of personally identifiable information (PII) such as names, detailed locations, email addresses, mobile numbers, etc., is thoroughly checked and removed.

This dataset includes 20 categories of jobs and CVs targeting these jobs. The categories are: \textit{Information Technology, Business Development, Finance, Advocate, Accountant, Engineering, Chef, Aviation, Fitness, Sales, Banking, Healthcare, Consultant, Construction, Public Relations, Human Resources, Designer, Arts, Teacher, Apparel}. In CVs, there are 10 for each category, but in job descriptions, there are more jobs in IT and engineering. 

Prompts are provided in Section \ref{sec:prompts}. While the core design remains the same, the prompts are slightly different for Curriculum Vitae and Job Description, each tuned to its specific modality. The full inference code with examples is available in the GitHub repository.

\subsection{Visualizing Knowledge Graphs}
{Utilizing the \textit{Gemini} tool, we systematically gathered data and constructed two knowledge graphs for CVs and job descriptions (JDs) as HR knowledge bases, adhering to the defined methodology. To ensure relevance, entities exceeding a length of 4 were excluded. These knowledge graphs (presented in Figures \ref{fig:KG_Examples/CV_mega} and  \ref{fig:KG_Examples/JD_mega}) show that there is a huge connection between different jobs and the skills, education, and experience required. By utilizing these relationships, many downstream tasks can be done. 

\begin{figure}[t] 
\centering {\includegraphics[scale=0.22]{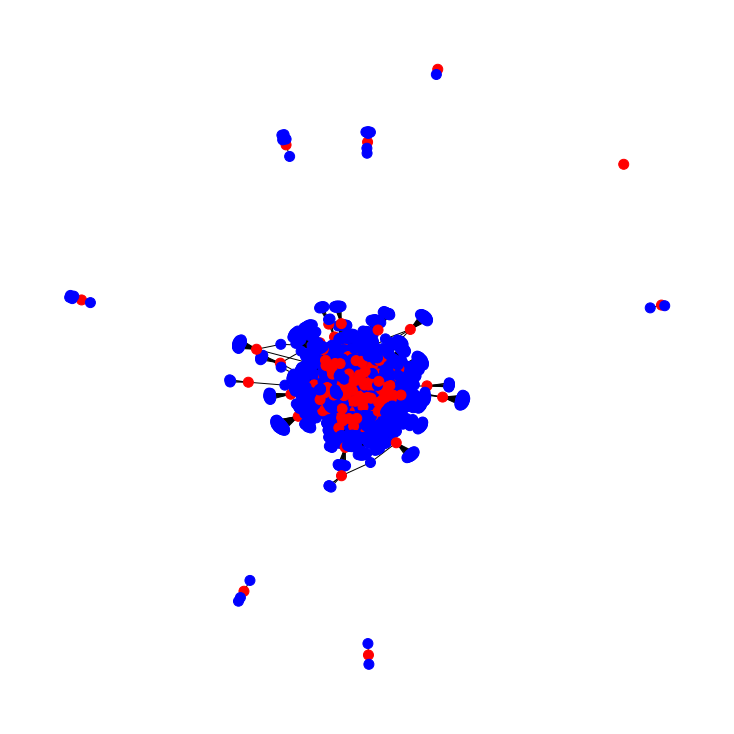}}
\vspace{-4mm}
\caption{Our Job Description Knowledge Graph. Red dots are Jobs, blue dots are skills, education and other entities}\label{fig:KG_Examples/JD_mega}
\vspace{-4mm}
\end{figure}

\section{Downstream Tasks}
\subsection{Information Propagation on HRGraph}
Figure \ref{fig:what-m} provides an overview of our Job Matching and Recommendation Framework utilizing HR Knowledge Graphs for job applicants or employees. Leveraging the job description graph, we identify matching skill, education, and experience nodes, forming a targeted sub-graph with 3-hop neighbouring nodes. Node centrality within this sub-graph allows us to efficiently find and rank all relevant job nodes.

\subsection{Task 01: Recommendation} \label{sec:DownstreamTask01-JobRecommendation}
 The information propagation framework predicts the top $N$ ranked jobs for each individual, enabling us to assess prediction accuracy and precision, thereby optimizing job recommendations. Similarly, in the task described above, we extend the methodology from Job Recommendation, utilizing the CV Knowledge Graph to identify employees based on matching skill, education, and experience. $R$ denotes random recommendations, $D$ denotes job recommendation using LLM entities directly for the top 5 recommendations. Table \ref{table:JobARec} shows that knowledge graph information propagation and ranking can provide very strong recommendations with good accuracy. { Case studies are provided in Appendix \ref{sec:app:exp-det}.}

\begin{table}[t]
\centering
\caption{Recommendation Results ($\uparrow$)}\label{table:JobARec}
    \begin{tabular}{c c c c }
        \hline
     $N$  &  \textbf{Task} & \textbf{Avg. Acc.} & \textbf{Avg. Prec.} \\
        \hline
       2 & Job Rec. & 0.668 & 0.675 \\
       2 & Employee Rec. & 0.684 &  0.685  \\
        \hline
       5 & Job Rec. & 0.748 & 0.764 \\
       5 & Employee Rec. & 0.784 &  0.792  \\
        \hline
       10 & Job Rec. & 0.702 & 0.700 \\
       10 & Employee Rec. & 0.715 &  0.708  \\
        \hline
       $D$ & Job Rec. & 0.670 & 0.655 \\
       $D$ & Employee Rec. & 0.620 &  0.665  \\
        \hline
       $R$ & Job Rec. & 0.323 & 0.312 \\
       $R$ & Employee Rec. & 0.373 &  0.361  \\
        \hline
    \end{tabular}
    \vspace{-3mm}
\end{table}

\subsection{Task 02: Job Area Classification}
In this task, we use KG-based job area classification on the CVs using two basic popular GNNs: GCN \citep{kipf2017semisupervised} and GAT \citep{veličković2018graph}. We compared the results with traditional and normally used deep learning models. Table \ref{table:JobAreaClassification} shows that Knowledge Graph-based GNN models are equally effective and slightly better than other models. {More details are provided in Appendix \ref{sec:app:exp-det}.}

\begin{table}[h]
\caption{Job Area Classification Results ($\uparrow$)}\label{table:JobAreaClassification}
    \begin{tabular}{c c c c }
        \hline
        \textbf{Model} & \textbf{Accuracy} & \textbf{Precision} & \textbf{Recall}\\
        \hline
        Tfidf+LogR. & 0.745 & 0.770 & 0.740\\
        Tfidf+DecT. & 0.655 &  0.670 & 0.655 \\
        Tfidf+RF & 0.680 &  0.675 & 0.680 \\
        Tfidf+GBC & 0.775 &  0.805 & 0.775 \\
        \hline
        Tfidf+MLP & 0.655 &  0.670 & 0.655 \\
        Transformer & 0.660 &  0.645 & 0.675 \\
        \hline
        GCN & 0.785 &  0.800 & 0.795 \\
        GAT & 0.775 &  0.835 & 0.775 \\
        \hline
    \end{tabular}
    \vspace{-5mm}
\end{table}

\section{Discussion}
We believe that transforming HR data into a knowledge graph holds a great promise in shaping the future of human resources data collection, management, and utilization. By envisioning HR data in this interconnected graph, organizations can unlock unprecedented insights, streamline recruitment processes, identify talent gaps, and foster employee growth. This approach not only enhances decision-making but also paves the way for a dynamic and adaptive HR ecosystem that propels organizational success in an ever-evolving landscape.

\section{Conclusion}
This study introduces a framework leveraging LLMs and GNNs to construct HR knowledge graphs from documents, working as a HR knowledge base for different HR tasks. The resulting KGs enhance various HR functions, including job matching, job area classification, and many more, demonstrating their efficacy through empirical evidence, benefiting both employers and employees.

\clearpage
\newpage

\section*{Limitations} 
The framework's primary limitation lies in its dependence on LLMs, which, although powerful, can be unreliable and prone to hallucinations \citep{wang2023assessing}. Given our model's exclusive reliance on LLMs for entity extraction, we observed instances where they deviated from the provided instructions. Also, a more sophisticated job-matching algorithm can be designed. Further research can be conducted on this to examine it in the future.

\section*{Ethical Considerations}
In conducting this research, strict ethical guidelines were followed to ensure the privacy and confidentiality of the individuals whose data was used. The primary focus was on handling personally identifiable information (PII) with the utmost care to protect the identity and privacy of all individuals.

\textbf{Data Anonymization.}
To safeguard privacy, all PII such as names, detailed locations, email addresses, and mobile numbers were meticulously identified and removed from the dataset. This process involved thorough checks to ensure no traceable information was left that could potentially identify any individual.

\textbf{Consent and Permissions.}
The original dataset was accessed with proper permissions and in compliance with the relevant data use agreements. By adhering to these agreements, we ensured that the data was used within the scope of its intended purpose, respecting the conditions under which the data was collected.

\textbf{Privacy Protection.}
In this research, the dataset utilized was curated from an existing collection of CVs, ensuring that all personally identifiable information (PII) was meticulously removed to maintain privacy and adhere to ethical guidelines. The original data sources were accessed with proper permissions, and stringent anonymization techniques were applied to eliminate any traces of identity. 

\textbf{Secure Data Handling.}
Throughout the data curation and analysis process, secure data handling practices were implemented. This included using encrypted storage solutions and restricting access to the data to only those team members who required it for their specific research tasks. These measures were crucial in preventing unauthorized access and potential data breaches.

\textbf{Ethical Use of Data.}
The research team was committed to using the data ethically, ensuring that the analysis and interpretations were fair and unbiased. The data was used solely for the purposes of this research and not for any commercial or exploitative activities. Additionally, findings were reported in a way that protected the anonymity of the individuals in the dataset.

\textbf{Transparency and Accountability}
Transparency in our methods and accountability in our processes were maintained throughout the research. Detailed documentation of our data handling and anonymization procedures was kept, ensuring that the steps taken to protect privacy could be reviewed and verified by external parties if necessary.

\section*{Acknowledgements}
I would like to express my sincere gratitude to \href{https://aclanthology.org/people/n/nikita-bhutani/}{Nikita Bhutani} for her immense support of my ideas and her extraordinary efforts as an advisor in finding valuable resources for the study. I also extend my thanks to \href{https://aclanthology.org/people/e/estevam-hruschka/}{Estevam Hruschka} for his insightful reviews and feedback. Additionally, I am particularly grateful to the \href{https://aclanthology.org/volumes/2024.nlp4hr-1/}{NLP4HR Workshop} \cite{nlp4hr-2024-natural} at EACL 2024 for providing me with the opportunities to be mentored and to develop my work.


\bibliography{custom}

\appendix

\section{Experimental Details}\label{sec:app:exp-det}
\subsection{Implementation Details}
\textbf{TF-IDF Vectorizer:} The model employs a TF-IDF Vectorizer with an n-gram range of 1 to 5, capturing diverse word combinations, and a maximum feature limit set to a calculated vocabulary size. The vocabulary size, determined as the mean plus three times the standard deviation of the data, ensures a comprehensive representation of relevant terms. Additionally, English stopwords are excluded to focus on meaningful content during the vectorization process.

\textbf{Traditional Models: } After getting vectors from TF-IDF Vectorizers, we use different methods to classify. \textit{LogR.} means \textit{Logistic Regression}, \textit{DecT.} means \textit{Decision Tree}, \textit{RF} means \textit{Random Forest} and \textit{GBC} denotes \textit{Gradient Boosting Classifier}.  Logistic Regression employs L1 regularization with the 'liblinear' solver. The Decision Tree Classifier has a maximum depth limited to 5. The RandomForest Classifier consists of 50 decision trees and uses a fixed random state for reproducibility. The Gradient Boosting Classifier incorporates an ensemble of 50 weak learners.

\textbf{MLP:} The MLP model is a simple feedforward neural network with multiple hidden layers, including dropout regularization for each layer. It consists of fully connected layers with decreasing dimensions from 2048 to 64 (halved in each layer), all utilizing the ReLU activation function. The output layer employs the softmax activation function for multi-class classification. 

\textbf{Transformer :} The transformer model, integrated with AutoML and the Hugging Face Transformers library, utilizes the AutoTokenizer to preprocess text data. The \textit{AutoModelForSequenceClassification} class is employed with the \textit{distilbert-base-uncased} model, configured to handle sequence classification tasks with a number of unique labels corresponding to the classes in the training data.

\textbf{Graph Neural Network-Based Models:} Both GCN \citep{kipf2017semisupervised} and GAT \citep{veličković2018graph} model used are the default models from \href{https://pytorch-geometric.readthedocs.io/}{Pytorch Geometric} library, with 64 hidden channels and 4 layers. Fine-tuning GNNs will improve the results.

\subsection{Case Study} 
In CV No. 92, it is a salesperson's CV. It has these matching entities with the job description graph: \textit{'accounting', 'managerial', 'excel', 'office', 'outlook', 'microsoft word', 'policies', 'sales', 'sap', 'time management'}. The top 5 matches job descriptions were: 150, 84, 103, 123, 163. The labels on them are
ACCOUNTANT, SALES, SALES, FINANCE, Sales respectively. While the individual's primary expertise lies in sales, the inclusion of the 'accounting' skill prompted a recommendation for an accountant role. Additional skills such as 'managerial,' 'excel,' and 'policies' contributed to suggestions within the finance industry. This exemplifies the Knowledge Graph's ability to provide nuanced explanations for recommendations, offering insights into the diverse factors influencing job suggestions. It can be very effective to make career-switch moves for job-seekers.

\section {Prompts} \label{sec:prompts}
If the information is a CV, use the following prompt:
\begin{quote}
You are an entity extraction expert, you can identify and extract different types of entities from a text. Here is some information from a CV. Your task is to find and enlist all the information entities like education (degree, grade, school name), skills (which skills the person has), qualifications (skills), experience (action verb and nouns), and any other helpful token that is important for a job, and share them in a list where entities are separated by commas. Do not write anything else. Just the small entities separated by commas in a dictionary (JSON). Each entity can have only 1-2 words.
\begin{verbatim}
<Insert CV text here>
\end{verbatim}
\end{quote}

If the information is a job description, use the following prompt:
\begin{quote}
You are an entity extraction expert, you can identify and extract different types of entities from a text. Here is some information from a job description. Your task is to find and enlist all the information entities like education (degree requirement), skills (which skills the job needs), qualifications (skills), experience (action verb and nouns), and any other helpful token that is important for a job, and share them in a list where entities are separated by commas. Do not write anything else. Just the small entities separated by commas in a dictionary (JSON). Each entity can have only 1-2 words.
\begin{verbatim}
<Insert job description text here>
\end{verbatim}
\end{quote}

Here is an example of the expected output:
\begin{quote}
{
    "Education": ["ABC University", "CGPA 3.00", "Computer Science and Engineering", "BSc"],
    "Skills": ["C", "Python", "R", "Machine Learning", "Communication", "Team Work"],
    "Experience": {
        "ABX InfoTech": ["Team Management", "Assistant Manager"],
        "STech": ["Manager", "Senior Engineer", "AWS"]
    }
}
\end{quote}

\end{document}